\documentclass[review]{elsarticle}

\usepackage{float}
\usepackage{graphicx}
\usepackage{multirow}
\usepackage{listings}
\usepackage{array}
\usepackage{lineno,hyperref}
\modulolinenumbers[5]
\usepackage{subcaption}
\usepackage{xcolor}
\usepackage{colortbl}
\usepackage{changepage}
\usepackage{tablefootnote}
\usepackage{caption}
\usepackage{mathtools}
\usepackage{amsfonts,amssymb,amsmath}

\journal{Journal of \LaTeX\ Templates}

\begin{document}

\begin{frontmatter}

% Change title from previous submission
%\title{TITLE}


\title{Merged ChemProt-DrugProt for Relation Extraction from Biomedical Literature}
%\author{
%  Shibani Likhite\textsuperscript{1}\thanks{slikhite@ucsd.edu} \and
% Jiawei Tang\and
%  Darshini Mahendran \and
%  Bridget T. McInnes  \and   Mai Nguyen\textsuperscript{2}\thanks{mhnguyen@ucsd.edu} 
%}

%\author{Clint Cuffy~\textsuperscript{1}, Logan Mills~\textsuperscript{2} and Bridget T. %McInnes~\textsuperscript{1}}
%\address{Virginia Commonwealth University, 401 S. Main St., Richmond, VA 23284, USA~\textsuperscript{1}}
%\address{University of Virginia, 1827 University Ave., Charlottesville, VA 22903, USA~\textsuperscript{2}}

%\ead{\{cuffyca, btmcinnes\}@vcu.edu}

\author{Mai H. Nguyen\textsuperscript{1*}, Shibani Likhite\textsuperscript{2}, Jiawei Tang\textsuperscript{2}, \\Darshini Mahendran\textsuperscript{3}, and Bridget T. McInnes\textsuperscript{3*}}

\address{San Diego Supercomputer Center, University of California San Diego~\textsuperscript{1}}
\address{Department of Computer Science \& Engineering, University of California San Diego~\textsuperscript{2}}
\address{Department of Computer Science, Virginia Commonwealth University~\textsuperscript{3}}

\cortext[mycorrespondingauthor]{Corresponding authors:  mhnguyen@ucsd.edu, btmcinnes@vcu.edu}
% \ead{mhnguyen@ucsd.edu,btmcinnes@vcu.edu}

\begin{abstract}

The extraction of chemical-gene relations plays a pivotal role in understanding the intricate interactions between chemical compounds and genes, with significant implications for drug discovery, disease understanding, and biomedical research.  This paper presents a data set created by merging the ChemProt and DrugProt datasets to augment sample counts and improve model accuracy. We evaluate the merged dataset using two state of the art relationship extraction algorithms: Bidirectional Encoder Representations from Transformers (BERT) specifically BioBERT, and Graph Convolutional Networks (GCNs) combined with BioBERT. While BioBERT excels at capturing local contexts, it may benefit from incorporating global information essential for understanding chemical-gene interactions. This can be achieved by integrating GCNs with BioBERT to harness both global and local context. Our results show that by integrating the ChemProt and DrugProt datasets, we demonstrated significant improvements in model performance, particularly in CPR groups shared between the datasets. Incorporating the global context using GCN can help increase the overall precision and recall in some of the CPR groups over using just BioBERT.

% The extraction of chemical-gene relations plays a pivotal role in understanding the intricate interactions between chemical compounds and genes, with significant implications for drug discovery, disease understanding, and biomedical research. In this paper, we present a comprehensive benchmarking study of Graph Convolutional Networks (GCNs) combined with Bidirectional Encoder Representations from Transformers (BERT) on our proposed merged ChemProt-DrugProt dataset, which comprises 66,000 chemical-gene relation pairs across ten relation classes. We conduct an ablation study to determine the most effective hyperparameter combinations for enhancing model performance. While BERT excels at capturing local context, it may benefit from incorporating global information essential for understanding chemical-gene interactions. This can be achieved by integrating GCNs with BERT to harness both global and local context. After evaluating the model using a combination of hyperparameter tuning techniques, we have concluded that masking target entities with their semantic types and downsampling of negative samples gave the highest accuracy of 89.33\% on the merged ChemProt-DrugProt dataset. Our results shed light on the strengths and weaknesses of this architecture within the context of our dataset.

\end{abstract}


\end{frontmatter}

% \linenumbers

\section{Introduction}

Biomedical literature serves as a vital conduit for various stakeholders within the scientific community, including biomedical researchers, clinicians, and database curators. Through articles, patents, and reports, these individuals disseminate their findings, contributing to the collective knowledge base of the field. However, the sheer volume of literature generated on a daily basis presents a significant challenge, hindering users' ability to efficiently retrieve relevant information. Consequently, there is a pressing demand for innovative solutions to streamline information retrieval processes. Recognizing this need, Natural Language Processing (NLP) systems have emerged as promising tools to automate the extraction of pertinent information from biomedical texts. By leveraging computational algorithms, NLP systems aim to expedite the identification and extraction of key insights, thereby alleviating the manual burden placed on users. NLP, as a field of study, is dedicated to developing techniques that enable computers to understand and analyze human language in its unstructured form. Within the realm of NLP, Relation Extraction (RE) stands out as a crucial area of focus. RE involves the identification and characterization of relationships between entities mentioned within textual data. By discerning connections between various entities—such as chemicals, genes, proteins, diseases, and treatments—RE facilitates the extraction of meaningful insights from biomedical literature. This capability holds immense potential for advancing biomedical research, clinical practice, and data curation efforts, ultimately driving innovation and improving outcomes within the healthcare domain.

In the landscape of biomedical literature analysis, a significant portion of existing systems is dedicated to the automatic recognition of mentions pertaining to genes, proteins, and chemicals within textual data.
While these systems play a pivotal role in facilitating the identification of individual entities, a noticeable gap exists in their ability to extract and elucidate the intricate interactions between these entities. 
Indeed, a limited number of approaches have been developed to specifically target the extraction of interactions between genes/proteins and chemicals within textual data. Given this context, there is a clear imperative to delve deeper into the diverse relationships that exist between drugs, chemical compounds, and various biomedical entities, particularly genes and proteins. Systematic extraction of these relationships is essential for enabling comprehensive analysis and exploration of key biomedical properties across a spectrum of applications. By honing in on these interactions, researchers can unravel crucial insights into drug mechanisms, disease pathways, and therapeutic targets, paving the way for advancements in drug discovery, personalized medicine, and disease management. Efforts to enhance the extraction of such relationships hold immense promise for bolstering the efficacy and efficiency of biomedical research and clinical practice. By leveraging advanced NLP techniques and innovative methodologies, researchers can unlock the full potential of biomedical literature, accelerating the pace of scientific discovery and revolutionizing healthcare delivery.

In this work, we focus on the creation of a unified dataset for chemical-gene relation extraction by merging two widely used resources: ChemProt and DrugProt. Although chemical-gene relation extraction has been extensively studied using a variety of deep learning methods—including RNNs~\cite{schmidt2019recurrent}, LSTMs~\cite{miwa2016end}, and CNNs~\cite{liu2013convolution}—our aim is to support future research by providing a merged, standardized benchmark dataset that incorporates both local (sentence-level) and global (document-level) contextual information. We evaluate the utility of this dataset by benchmarking two established approaches: a BERT-based model~\cite{devlin2018bert} for local context modeling, and GCN-BERT~\cite{mahendran2022graph}, which augments BERT with graph convolutional networks to integrate global context across documents.

Section~\ref{sec:data} describes the dataset merging strategy. We assess the quality and effectiveness of the merged dataset using the BERT model in Section~\ref{sec:biobert}, and evaluate it further with GCN-BERT in Section~\ref{sec:gcn-bert}. Experimental setup and results are presented in Sections~\ref{sec:exp_setup} and \ref{sec:results}, providing a comparative view of model performance on the new dataset. Section~\ref{sec:discussion} discusses the implications of our findings, and we outline directions for future dataset development and evaluation strategies in Section~\ref{sec:future-work}.

\section{Related work}

Many approaches have been explored for RE in the biomedical domain.  These approaches can be divided into 4 major classes: 1) rule-based, 2) machine learning-based 3) deep learning-based, and 4) contextualized language model-based approaches.

Rule-based approaches utilize rules and patterns to identify relations between words. He, et al.~\cite{he2020extended} proposed a rule-based method to identify relations between chemical reactions synthesis by compiling two dictionaries, first dictionary they used to identify the entities, and the second to identify the relations between entities that occur in the same sentence. Li, et al.~\cite{li2015end} proposed a rule-based method where they used regular expressions to match drug names to a prescription list then co-location information to link the attributes to the drugs. Lowe, et al.~\cite{loweextraction} used the ChemicalTagger~\cite{hawizy2011chemicaltagger} an open-source tool that uses syntactic information to identify relations between entities.

Machine learning-based approaches use machine learning algorithms and statistical analysis to identify traditional features best able to classify relations. SVMs dominated most of the machine-learning-based approaches~\cite{patrick2010i2b2,grouin2010caramba,demner2010nlm,solt2010concept}. Miller, et al.~\cite{miller2019extracting} used token context features, character type features, and semantic features with SVMs. Anick, et al.~\cite{anick2010i2b2} used lists of n-grams with specific semantics as the features for SVM. Demner-Fushman, et al.~\cite{demner2010nlm} incorporated semantic information using concepts from the UMLS~\cite{aronson2001effective} and exercised feature reduction through cross-validation. Zhu, et al.~\cite{zhu2013detecting} focused on revising the learning algorithm by reformulating the SVM into a composite-kernel framework to achieve better performance.

Hybrid approaches combined linguistic pattern matching with ML techniques. Grouin, et al.~\cite{grouin2010caramba} and Minard, et al.~\cite{minard2011hybrid} also trained an SVM and constructed linguistic patterns manually. They reported that there are advantages of the hybrid approach as linguistic patterns may confirm automatically-induced relations which helps adding confidence to the obtained results. While Yang, et al.~\cite{yang2020identifying} applied heuristic rules to generate candidate pairs of possible related entities that were fed into three ML models (SVM, RF, and Gradient Boosting~\cite{friedman2002stochastic}) to classify the relations. The possible related entities were identified according to their distance from each other as defined by the number of sentence boundaries between the two entities, and multiple classifiers were developed to classify relations based on this distance.

Deep learning is a field derived from machine learning. Sahu, et al.~\cite{sahu2016relation} and Luo, et al.~\cite{luo2017segment} utilized convolutional neural network (CNN) to automatically learn feature representations to reduce the need for engineered features. Sahu, et al.~\cite{sahu2016relation} applied CNN to build a Sentence-CNN which learns a single sentence-level representation for each relation, using discrete features. Lv, et al.~\cite{lv2016clinical} proposed the adoption of a CRF model and applied a deep learning model for features optimization by the employment of autoencoder and sparsity limitation. Tang, et al.~\cite{tang2019convolutional} used a hierarchical attention-based convolutional LSTM (ConvLSTM) model to construct a sentence as a multi-dimensional hierarchical sequence, to learn the local and global context information. Wei, et al.~\cite{wei2020study} proposed a model combining CNN and RNN where they utilized local context and semantic features as features~\cite{patrick2010high}.

Leveraging contextualized language models for RE in the biomedical domain has been gaining attention. Alimova, et al.~\cite{alimova2020multiple} conducted a comparison between three BERT-based models: BERT-uncased~\cite{devlin2018bert}, BioBERT~\cite{lee2020biobert}, and Clinical BERT~\cite{alsentzer2019publicly} for extracting relations from clinical texts. Copara, et al~\cite{copara2020named} used a BERT-based method assessing five variations of the BERT language models, including a domain-specific model called ChemBERTa.  
Zhang, et al.~\cite{zhangmelaxtech} proposed a hybrid method combining deep learning models with pattern-based rules and built a binary classifier by fine-tuning BioBERT.
In previous works, we utilized two general BERT and a BioBERT models to automatically detect relations between chemical compounds/drugs and genes/proteins~\cite{mahendran2021biocreative}, and   
explored rule-based, deep learning based and BERT-based methods to identify adverse drug events (ADEs) from clinical text~\cite{mahendran2021extracting}.

Several groups have explored combining BERT-based models with graph convolutional networks (GCNs) for biomedical NLP tasks. 
Lai \& Lu~\cite{lai2021bert} presented an approach that combine BERT and a graph transformer with neighbor attention to capture relations across sentence boundaries. Our previous work proposed combining BERT with Graph Convolutional Neural Networks to extract relationships between chemicals reactions and their parameters from chemical patents~\cite{mahendran2022graph}. Subsequently, a number of other systems have been proposed. For example, Nojoo Kambar et al.~\cite{nojoo2022chemical} proposed a model that integrates a BERT-based encoder with a GCN for joint entity and relation extraction of Genes and Chemicals from biomedical texts.  While their model is specific to Gene and Chemicals, our previous model is more generalizable across a variety of different entity types. 

Zhou, et al.~\cite{zhou2023llm} utilize large language models (LLMs) by introducing  LEAP, a framework using adaptive instructions and examples to find relationships in clinical data. Yoon, et al.~\cite{yoon2023biomedical} proposed a system that uses LLMs, a combination of weakly labeled data and knowledge bases to achieve better performance than standard methods. Yuan, et al.~\cite{yuan2021improving} created a new model, KeBioLM, that uses a well-established knowledge base called UMLS to understand biomedical text by incorporating existing medical knowledge. Tinn, et al.~\cite{tinn2023fine} and Delmas, et al.~\cite{delmas2023relation} explored training with limited data. Tinn, et al. used a medical vocabulary and specialized pre-training to create robust models for biomedical applications whereas Delmas, et al. explored different techniques including training with synthetic data generated by LLMs. Dunn, et al.~\cite{dunn2022structured} presented a simple sequence-to-sequence approach to joint NER and RE for complex hierarchical information by leveraging a fine-tuned LLM.

%%%%%%%%%% DATA %%%%%%%%%%
\section{Data}
\label{sec:data}

In this section, we describe the ChemProt and DrugProt datasets, our merging strategy and resulting new, larger and more comprehensive dataset. 

\subsection{Datasets}

The ChemProt and DrugProt datasets, both released as part of the BioCreative challenge~\footnote{https://biocreative.bioinformatics.udel.edu/}, contain chemical-protein/gene interactions and provide manual annotations for these entities along with their relations within their respective PubMed abstracts. There are differences between the two datasets in terms of the number of abstracts, entities, and relations. 
% The ChemProt training data contains 1,020 abstracts with 25,752 entities and 6,437 relations; and the ChemProt validation data contains 612 abstracts with 15,567 entities and 3,558 relations. The DrugProt training data contains 3,500 abstracts with 89,529 entities and 17,274 relations; and the DrugProt validation data contains 750 abstracts with 18,858 entities and 3,765 relations. 
Notably, due to its later release, DrugProt contains more chemical-gene entity relations compared to ChemProt, as indicated in Table~\ref{tab:data}. 

\begin{table}[h]
\centering
\caption{ChemProt vs. DrugProt Datasets\\}
% \scalebox{0.8}{
\resizebox{0.8\textwidth}{!}{
\begin{tabular}{l|r|r|r|r}
\hline
% \toprule
& \multicolumn{2}{|c}{Train Data} & \multicolumn{2}{|c}{Validation Data} \\
& ChemProt & DrugProt & ChemProt & DrugProt \\
\hline
% \multirow{3}{*}{cellline}
Number of Abstracts & 1,020 & 3,500 & 612 & 750 \\
Number of Entities & 25,752 & 89,529 & 15,567 & 18,858 \\
Number of Relations & 6,437 & 17,274 & 3,558 & 3,765 \\
% \bottomrule
\hline
\end{tabular}
}
\label{tab:data}
\end{table}

\begin{figure*}[t]  % Use figure* to span both columns
    \centering
    \includegraphics[width=\textwidth]{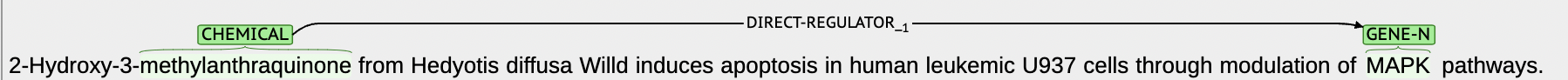}  % Replace with your image file name and extension
    \caption{Illustration of entities and relations in from sample input sentences}
\label{fig:input_example}
\end{figure*}

Figure \ref{fig:input_example} illustrates sample  sentences with annotated entities and their relations.

\begin{figure*}[t]  % Use figure* to span both columns
    \centering
    \includegraphics[width=\textwidth]{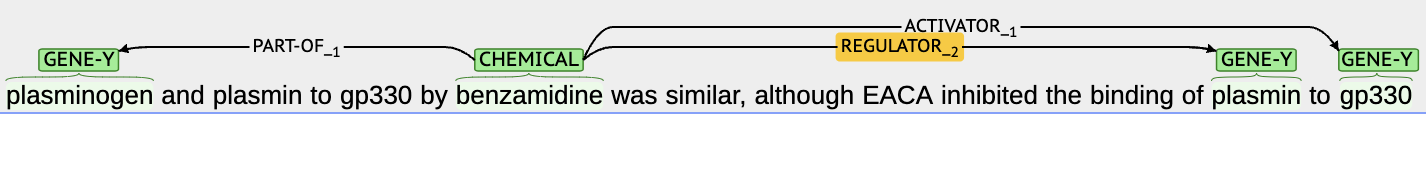}  % Replace with your image file name and extension
\end{figure*}

Both ChemProt and DrugProt classify their chemical-gene relations into 22 categories, further organized into 10 ChemProt Relation (CPR) groups. The definitions of these groups, as outlined in ChemProt's documentation, guided our preprocessing efforts. We systematically mapped all 22 relation categories to these 10 groups, as shown below, to facilitate a more streamlined analysis:
{\footnotesize 
\begin{itemize}
    \setlength{\itemsep}{-2pt}
    % \setlength{\leftmargin}{-6pt}
    % \small
    \item CPR:1  - PART\_OF   
    \item CPR:2  - REGULATOR, DIRECT\_REGULATOR, INDIRECT\_REGULATOR   
    \item CPR:3  - UPREGULATOR, ACTIVATOR, INDIRECT\_UPREGULATOR   
    \item CPR:4  - DOWNREGULATOR, INHIBITOR, INDIRECT\_DOWNREGULATOR   
    \item CPR:5  - AGONIST, AGONIST\_ACTIVATOR, AGONIST\_INHIBITOR   
    \item CPR:6  - ANTAGONIST  
    \item CPR:7  - MODULATOR, MODULATOR\_ACTIVATOR, MODULATOR\_INHIBITOR  
    \item CPR:8  - COFACTOR  
    \item CPR:9  - SUBSTRATE, PRODUCT\_OF, SUBSTRATE\_PRODUCT\_OF  
    \item CPR:10 - NO RELATION 
\end{itemize}
}

%\begin{table}[H]
%\centering
%\resizebox{1.0\textwidth}{!}{
%\large
%\renewcommand{\arraystretch}{1.5} 
%\large
%\begin{tabular}{|c|c|c|c|c|}
%\hline
%& \multicolumn{2}{|c|}{Train Data} & \multicolumn{2}{|c|}{Validation Data} \\
%\hline 
%& ChemProt & DrugProt & ChemProt & DrugProt \\
%\hline
%Number of Abstracts & 1,020 & 3,500 & 612 & 750 \\
%\hline
%Number of Entities & 25,752 & 89,529 & 15,567 & 18,858 \\
%\hline
%Number of Relations & 6,437 & 17,274 & 3,558 & 3,765 \\
%\hline
%\end{tabular}
%}
%\caption{ChemProt vs. DrugProt Datasets}
%\label{tab:dataset-diff}
%\end{table}

%\begin{table}
%\centering
%\resizebox{1.0\textwidth}{!}{
%\large
%\renewcommand{\arraystretch}{1.5} 
%\begin{tabular}{|l|l|} \hline  % Adjust column width as needed
%CPR Group & Relations \\ \hline 
%CPR:1  & PART\_OF \\ \hline  
%CPR:2  & REGULATOR, DIRECT\_REGULATOR, INDIRECT\_REGULATOR \\ \hline  
%CPR:3  & UPREGULATOR, ACTIVATOR, INDIRECT\_UPREGULATOR \\ \hline  
%CPR:4  & DOWNREGULATOR, INHIBITOR, INDIRECT\_DOWNREGULATOR \\ \hline  
%CPR:5  & AGONIST, AGONIST\_ACTIVATOR, AGONIST\_INHIBITOR \\ \hline  
%CPR:6  & ANTAGONIST \\ \hline  
%CPR:7  & MODULATOR, MODULATOR\_ACTIVATOR, MODULATOR\_INHIBITOR \\ \hline  
%CPR:8  & COFACTOR \\ \hline  
%CPR:9  & SUBSTRATE, PRODUCT\_OF, SUBSTRATE\_PRODUCT\_OF \\ \hline 
%CPR:10 & NO RELATION \\ \hline
%\end{tabular}
%}
%\caption{CPR Group-to-Relations Mapping}
%\label{tab:data}
%\end{table}

\subsection{Dataset Merging Processing}

We developed a merging strategy to combine the ChemProt and DrugProt datasets due to the considerable overlap between the two. As DrugProt contains a larger number of relations than ChemProt, straightforward merging was challenging.

We initially organized entities and relations based on the abstracts in which they appear, resulting in distinct sets for ChemProt and DrugProt abstracts. During the merging process, abstracts exclusively present in one set are directly added to the merged set. In instances where an abstract exists in both sets, we merged entities and relations from both datasets for that abstract. For these common entities, no inconsistencies were detected in terms of text and entity content; however, relation conflicts were identified. In the training sets, 63 relation conflicts were identified, along with 7 conflicts in the validation sets. These conflicts were resolved manually. 
The size of the merged dataset and the individual ChemProt Relation (CPR) group sizes are detailed in Table \ref{tab:merged-dataset}.

\begin{table}[h]
\centering
% \scalebox{0.8}{
\resizebox{1.0\textwidth}{!}{
\begin{tabular}{l|r|r}
\hline
& Merged Train Data & Merged Validation Data \\
\hline
% \multirow{3}{*}{cellline}
Number of Abstracts & 3,824 & 1,184 \\
Number of Entities & 97,597 & 29,763 \\
Number of Relations & 20,401 & 6,450 \\
\hline
\hline
% \toprule
CPR Group & Training Data & Validation Data \\
\hline
% \multirow{3}{*}{cellline}
CPR:1 & 1,041 & 352 \\
CPR:2 & 3,463 & 1,183 \\
CPR:3 & 3,101 & 984 \\
CPR:4 & 7,453 & 2,217 \\
CPR:5 & 781 & 226 \\
CPR:6 & 1,045 & 368 \\
CPR:7 & 29 & 19 \\
CPR:8 & 32 & 2 \\
CPR:9 & 3,214 & 922 \\
CPR:10 & 262 & 178 \\
% \bottomrule
\hline
\end{tabular}
}
\caption{Merged ChemProt-DrugProt Dataset\\}
\label{tab:merged-dataset}
\end{table}

%While converting from JSON, we implemented a strategy to increase the number of samples, leading to a significant number of CPR-10 samples in the merged dataset. If a given chemical-gene pair did not have any annotated relation, they were assigned a CPR-10 (no relation) label. 
The BRAT rapid annotation tool~\cite{inproceedings} has been widely used in the NLP research community for various annotation tasks, including named entity recognition, relation extraction, and co-reference resolution. This annotation format is designed in particular for structured annotation, where the notes are not freeform text but have a fixed form that can be automatically processed and interpreted by a computer. All annotations follow the same basic structure: Each line contains one annotation, and each annotation is given an ID that appears first on the line, separated from the rest of the annotation by a single TAB character. The rest of the structure varies by annotation type. We converted the merged dataset from JSON to annotations in BRAT format, utilizing the .ann (annotated files) and .txt files for each abstract in the dataset. 

%The merging and preprocessing pipeline is represented in Figure \ref{fig:data_merging}.

%\begin{figure}[h]  % Use figure* to span both columns
%    \centering
%    \includegraphics[width=1\textwidth]{Figures/merging_data.png}  
%    \caption{Data Merging and Preprocessing pipeline}
%    \label{fig:data_merging}
%\end{figure}

\subsection{Training Data}
% To facilitate model training, we randomly split our original merged training set into an 80/20 ratio, creating separate training and validation sets. The original merged validation dataset was subsequently repurposed as the test set for final evaluation. During training, the 20\% portion of the original merged training data served as the validation set for hyperparameter tuning and model optimization.
The ChemProt and DrugProt datasets were part of the BioCreative challenge, thus only training and validation data were made available, i.e, the test data was not made available.  We created merged ChemProt-DrugProt training and validation datasets as described above.  For model training, the merged validation dataset was used as the held-out test set for final evaluation.  The merged train dataset was partitioned using an 80/20 ratio with stratified sampling to create train and validation datasets for model development and hyperparameter tuning.

% For our reported results, CPR groups 7 and 8 were omitted due to their insufficient representation, which poses challenges for model generalization. 
For our reported results, CPR groups 7 and 8 were omitted due to their insufficient representation.  Since the sample counts in these groups are so small compared to the other groups, the model may never learn these classes, leading to poor generalization.
Additionally, to augment the negative class (CPR:10), we introduced additional instances by selecting pairs of chemical and gene entities lacking any association in each input sentence. In other words, if a given chemical-gene pair did not have any annotated relation, they were assigned a CPR-10 (No Relation) label. This approach generated more explicit negative samples to prevent the model from making spurious relation predictions where there was none. 
% This approach contributed to the generation of more representative negative class instances.

%%%%%%%%%% BioBERT %%%%%%%%%%

% \section{BioBERT Approach}
\section{Relation Extraction Methods} 

\subsection{BioBERT Model Setup}
\label{sec:biobert}

Our BioBERT-based model comprises of two primary components: a BioBERT model and a fully connected top layer referred to as the "top model." BioBERT, developed by Lee et al.~\cite{lee2020biobert}, is a contextual embedding model built upon the architecture of BERT. This model has been pre-trained using a vast corpus of biomedical texts, including PubMed abstracts and full-text articles from PMC.  The top model is a multi-layer fully connected network that is stacked on top of BioBERT  and serves as the classification component in our RE model. The model begins by processing the embedding of the [CLS] token extracted from BioBERT. This embedded representation then passes through two hidden layers, each consisting of 1024 units, before reaching the output layer. To ensure compatibility with the training data, our output structure's dimensions are aligned with the count of CPR groups present in the training data. Each entry within the vector of this structure represents the probability of the input being associated with the corresponding CPR group. 

The BioBERT model was fine-tuned on the training dataset, followed by evaluation using the  validation dataset. To update the model parameters, we utilized the Adam optimizer with a weight decay of 0.01. We used a variable global learning rate scaling approach in which the learning rate increases linearly during the warmup stage, and is equal to the inverse squared-root of the step count after the warmup stage:
\[
lr = lr\_factor \cdot \min\left(step^{-0.5}, step \cdot warm\_up^{-1.5}\right)
\]
In our experiments, we use $lr\_factor=0.0005$ and $warm\_up=1000$.

Throughout the training process, we implemented an early stopping mechanism with a patience threshold set to 6 steps. Subsequently, the best-performing model, as determined by the early stopping process, underwent evaluation using the test dataset. All three models underwent a training regimen of 5 epochs and were trained 5 times with varied random initializations.

%%%%%%%%%% GCN-BERT %%%%%%%%%%

\subsection{GCN-BERT Model Setup}
\label{sec:gcn-bert}
With the effectiveness of our custom merged dataset established, we then proceeded to performing experiments with the GCN-BERT model to capture both local and global context for chemical-gene RE.  Global context is important for incorporating information from co-occurring words.  The GCN-BERT architecture is based on three key modules:  vocabulary graph, Graph Convolutional Network and the combined GCN-BERT. The specifics of these modules are given in the sub-sections below.

\subsubsection{Vocabulary Graph}
The abstract text undergoes initial preprocessing, which includes the removal of stop words, punctuations, etc. The sentences are then tokenized using the BioBERT tokenizer \cite{lee2020biobert}. The tokenizer maps these tokens to integers and handles out-of-vocabulary (OOV) words by splitting them into subwords and using '\#\#' to represent split tokens. This processed input is used to create a vocabulary graph.

The number of nodes within this graph corresponds to the number of unique tokens in the entire corpus. Word nodes in the graph are represented with assigned integers. We calculate the weight of the edge between two word nodes using Pointwise Mutual Information (PMI). PMI is a metric that quantifies the probability of two words co-occurring. To compute the PMI score, we create a map of word-pair occurrences within a window of size 20. The frequency of these pairs across all windows is calculated to determine the PMI score between two word tokens. As a result, edges are established between word nodes when the PMI value is positive during the graph generation process.
Therefore, if a particular chemical-gene pair appears multiple times throughout the corpus, that pair will have a positive PMI value. 

The PMI formula \cite{bouma2009normalized} is expressed as follows:
\begin{equation}
PMI(x, y) = \log\left(\frac{P(x, y)}{P(x) \cdot P(y)}\right)
\end{equation}

where:
\begin{equation}
P(x, y) = \frac{\#SW(x, y)}{\#SW} \text{ and }  P(x) = \frac{\#SW(x)}{\#SW}
\end{equation} 

\emph{\#SW(*)} represents the number of sliding windows containing word pair or a particular word, and \emph{\#SW} represents the number of sliding windows.

In this context, the PMI values are normalized, resulting in Normalized PMI (NPMI) values within the range of [-1, 1]. A positive NPMI value indicates a significant semantic association between words, whereas a negative NPMI value suggests little to no semantic connection. 

\begin{figure*}[t]  % Use figure* to span both columns
    \centering
    \includegraphics[width=\textwidth]{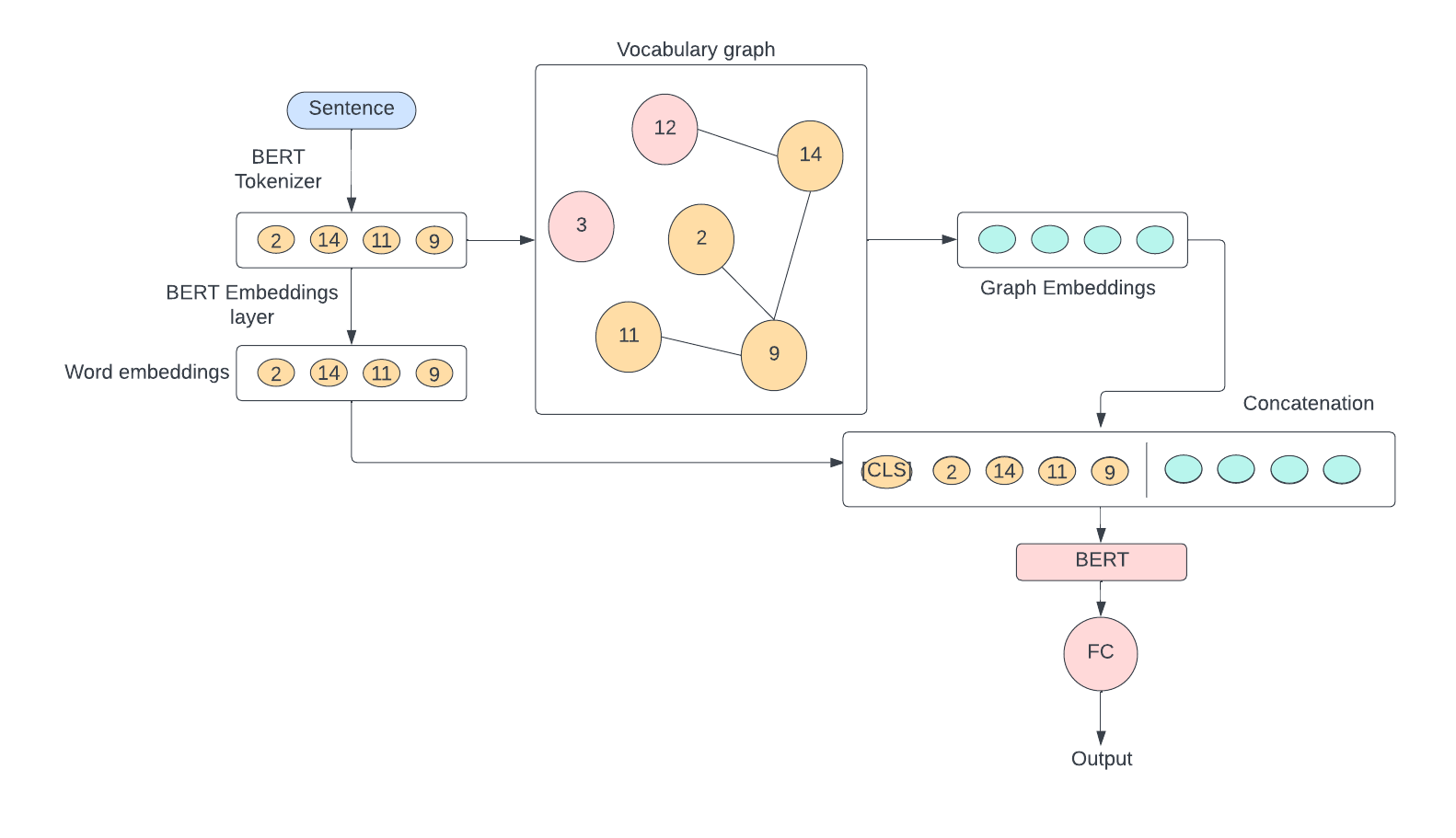}  % Replace with your image file name and extension
    \caption{Architecture of GCN-BERT adapted from \cite{mahendran2022graph}}
    \label{fig:architecture_diag}
\end{figure*}
\subsubsection{Graph Convolutional Network}
Graph Convolutional Networks (GCNs) serve as a fundamental tool in analyzing complex chemical-gene interactions in the GCN-BERT model. The vocabulary graph is passed through a two-layer GCN, aiming to create graph embeddings that leverage the connections between neighboring nodes. GCNs are particularly valuable in the context of non-Euclidean data, as they excel at capturing intricate relations among entities.

One of the key strengths of GCNs lies in their ability to perform message passing within the graph. This means that the model can gather information from neighboring nodes, allowing it to grasp the broader context of chemical-gene interactions. The number of GCN layers used in our approach directly impacts the depth of information gathering, determining the maximum number of hops a node can make to capture information even from nodes that are not directly connected.

Our vocabulary graph is represented as \emph{G (N, E)}, where \emph{N} is the set of nodes, with a node for each token in the corpus, and \emph{E} denotes the set of edges within the graph. The vocabulary graph serves as a structure through which information about the semantic relations among words is propagated. The GCN model performs convolutions over this graph, aggregating information from the nodes' neighborhoods to update their feature representations. This process results in the generation of enriched graph embeddings that capture both local and global semantic relations within the corpus.

A single convolutional layer of GCN can be calculated as:

\begin{equation}
H = \widetilde{A} XW
\end{equation}

%Rephrase, please do not copy paste this, slight rephrase for a plagiarism detection engine 
In this equation, $X \in \mathbb{R}^{n \times m}$ is the input matrix having 'n' nodes with 'm' dimensional features. $W \in \mathbb{R}^{m \times h}$  represents the weight matrix, and $\widetilde{A} = D^{-\frac{1}{2}}AD^{-\frac{1}{2}}$  defines the normalized symmetric adjacency matrix. Here $D$ is defined as $D_{ii} = \sum_{j} A_{ij}$. The $\widetilde{A}X$ term captures the part of the graph relevant to the input.

For a two-layer Vocabulary GCN with a ReLU activation function, the representation is as follows:

\begin{equation}
VGCN = \text{ReLU}(X_{mv} \widetilde{A}_{vv} W_{vh}) W_{hc}
\end{equation}

where 'm' represents the mini-batch size, 'c' is the size of the sentence embedding. 'v' is the vocabulary size and 'h' is the size of the hidden layer.

\subsubsection{Combined GCN-BERT}
A sentence containing entities is extracted from the abstract and fed to GCN-BERT. This sentence containing these target entities is processed as follows:

\begin{itemize}
    \item Tokenization: We first tokenize each input sentence using the BioBERT tokenizer.
\end{itemize}

\begin{itemize}
    \item Embeddings: The resulting tokenized sentence, containing a specific entity pair, is passed through the BioBERT embedding layer to generate embeddings for each token in the sentence.  
    
    \item Vocabulary Graph Embeddings: Simultaneously, the tokenized input sentence is processed through the Graph Convolutional Network (GCN) to extract graph embeddings (as indicated by the blue ovals in Figure \ref{fig:architecture_diag}) for the relevant part of the input, which represent relationships between entities.
    
    \item Concatenation: The graph embeddings are concatenated with the BioBERT embeddings, resulting in an ensembled embedding representation. 
    
    \item GCN-BERT: This combined embedding is then passed through the BERT model, which consists of a stack of 12 encoders in a feedforward neural network. This configuration is a standard choice in BERT models. The model is used for bidirectional training to produce the final vector representation.
    
    \item Classification: Finally, this vector is passed through a fully connected layer for the final classification task.
\end{itemize}

%%%%%%%%%% EXPERIMENTAL SETUP %%%%%%%%%%
\section{Experimental Setup}
\label{sec:exp_setup}

We explored a range of tuning techniques and combinations to optimize both the BERT and GCN-BERT model.

\subsection{Entity Masking}
Entity masking is a crucial technique for enhancing the model's ability to generalize relationships from text. We employed two entity masking strategies:

\begin{itemize}
    \item Masking Target Entities: Replacing the target entities with their entity type while leaving non-target entities unchanged, also known as reverse entity masking. This strategy resulted in the highest accuracy in our experiments. 
    \item Masking Non-Target Entities:  Masking the non-target entities from the sentence while keeping target entities intact.
\end{itemize}

\subsection{Downsampling Negative Samples}
The negative class, augmented with generated negative class instances as described in Section~\ref{sec:data}, ended up being much larger than the other classes.  
To address this class imbalance in our dataset, we implemented a 0.6 downsampling ratio for negative samples.
%, reducing their count from 13,849 to 9,168. 
This technique significantly improved the model's ability to handle class imbalance. 

\subsection{Weighted Loss}
We incorporated weighted loss during training to tackle class imbalance. Weighted loss assigns higher importance to minority classes, effectively addressing the imbalanced sample distribution. This technique was particularly effective given the prevalence of negative class samples in our dataset. The weighted loss can be calculated as:
\[
\text{Weighted Loss} = \sum_{i=1}^{N} w_i \cdot L(y_i, f(x_i))
\]
where \(N\) is the number of samples, \(w_i\) is the weight assigned to the i-th sample and \(L(y_i, f(x_i))\) is the loss for the i-th sample, typically the cross-entropy loss.

%\subsection{Training}
%We conducted experiments to optimize our model's performance by varying the batch size, learning rate, and the number of training epochs. Specifically, we employed a batch size of 8, the Adam optimizer with a learning rate of 1e-5, and 15 training epochs.

%We implemented the BioBERT model using PyTorch-Transformers, developed by the HuggingFace team \cite{lee2020biobert}. To initialize our model, we used the weights from the BioBERT-Base model, along with BioBERT word embeddings. For some experiments, BioBERT-Large weights were used instead to determine the effects of the larger model on overall performance.

% The model training process spanned approximately 10 hours and was carried out on an NVIDIA GeForce RTX 2080 Ti.

% \subsubsection{Using BioBERT-Large}
% In experiments, we evaluated the impact of model size by using BioBERT-Large weights instead of BioBERT-Base. BioBERT-Large is based on BERT-Large and features twice as many layers, more attention heads, and parameters compared to BioBERT-Base. Additionally, it is trained on a larger biomedical dataset. These experiments aimed to determine if the increased model size could improve overall performance.

%%%%%%%%%% RESULTS %%%%%%%%%%
%\section{Experimental Details}
\section{Results} 
\label{sec:results}

\subsection{Evaluation Metrics}

We assessed the model's performance using several evaluation metrics, including precision, recall, Micro F1, Macro F1, and Weighted F1 scores.

The Macro F1 score represents the average of F1 scores across all classes and can be calculated as follows:
\[
\text{Macro F1} = \frac{1}{N} \sum_{i=1}^{N} F1_i
\]
where \(N\) is the number of classes, and \(F1_i\) is the F1 score for class \(i\).

The Weighted F1 score is computed by taking the mean of per-class F1 scores while considering the support of each class. In this context, "support" refers to the proportion of each class's representation relative to the total number of instances. It can be calculated as:
\[
\text{Weighted F1} = \frac{1}{N} \sum_{i=1}^{N} \frac{N_i}{N} \times F1_i
\]
where \(N\) is the number of classes, \(N_i\) is the number of true samples in class \(i\), and \(F1_i\) is the F1 score for class \(i\).

The Micro F1 score, often referred to as accuracy, measures the proportion of correctly classified observations out of the total number of observations and can be calculated as:
\[
\text{Micro F1} = \frac{2 \times \text{TP}}{2 \times \text{TP} + \text{FP} + \text{FN}}
\]

\subsection{Merged Dataset Analysis}

We conducted a comprehensive comparison between the performance of our merged dataset and the ChemProt dataset processed by Sun et al.~\cite{sun2018hierarchical} For this evaluation, we specifically focused on CPR groups 3, 4, 9, and 10, as they are common to both datasets processed by Sun et al. The substantial increase in the size of our dataset led to a significant improvement in performance. The results for various CPR groups are presented in Table~\ref{tab:chemprot-vs-merged-results}, with Precision, Recall and $F_1$ scores averaged over five execution runs. Notably, our model demonstrates a marked increase in performance across all positive groups (CPR 3, 4, and 9), attributed to the augmented sample counts within these groups. By concentrating on CPR groups 3, 4, and 9, shared between the ChemProt and DrugProt datasets, we aim to underscore the effectiveness of our approach. These groups serve as a crucial benchmark for evaluating the impact of our merged dataset, representing interactions consistently captured across both datasets. Analyzing the performance enhancements in these shared groups highlights how increased sample counts resulting from the merging of ChemProt and DrugProt contribute to the overall effectiveness of our BioBERT model. This focused comparison demonstrates the tangible benefits of our methodology in accurately predicting chemical-gene interactions.

\begin{table}[h]
\centering
\caption{Results on ChemProt vs Merged Datasets with BioBERT-Base Model\\}
% \scalebox{0.8}{
\resizebox{0.8\textwidth}{!}{
\begin{tabular}{l|c|c|c|c|c|c}
\hline
% \top& rule
      & \multicolumn{3}{|c|}{ChemProt} & \multicolumn{3}{|c|}{ChemProt-DrugProt}  \\
\hline
      & P & R & F                          & P & R & F                         \\ 
\hline
CPR-3	& 0.7372	& 0.7407	& 0.7384	& 0.8690	& 0.8785	& 0.8735 \\
CPR-4	& 0.7951	& 0.8236	& 0.8083	& 0.8894	& 0.9263	& 0.9073 \\
CPR-9	& 0.6026	& 0.6976	& 0.6410	& 0.7620	& 0.8520	& 0.8040 \\
CPR-10	& 0.9544	& 0.9415	& 0.9479	& 0.9725	& 0.9575	& 0.9649 \\
\hline
\hline
Micro F1 & \multicolumn{3}{|c|}{0.7839} & \multicolumn{3}{|c}{0.8875} \\
Std dev  & \multicolumn{3}{|c|}{0.0025} & \multicolumn{3}{|c}{0.0021}  \\
\hline
\end{tabular}
}
\label{tab:chemprot-vs-merged-results}
\end{table}

% Repeat of what's above table -- By focusing on CPR groups 3, 4, and 9, which are common between the ChemProt and DrugProt datasets, we aim to highlight the effectiveness of our approach. These specific groups serve as a crucial benchmark for evaluating the impact of our merged dataset, as they represent interactions that are consistently captured across both datasets. By analyzing the performance improvements in these shared groups, we can demonstrate how the increased sample counts resulting from the merging of ChemProt and DrugProt contribute to enhancing the overall effectiveness of our relation extraction model. This targeted comparison allows us to showcase the tangible benefits of our methodology in capturing and accurately predicting chemical-gene interactions that are pertinent to both datasets, thus affirming the utility and robustness of our approach.

%%% BioBERT-Base vs BioBERT-Large
The BioBERT~\cite{lee2020biobert} model offers two configurations: 'Base' and 'Large.' BioBERT-Large's pre-training was performed on a more extensive vocabulary that encompasses a broader spectrum of biomedical terms compared to BioBERT Base. 
%The key distinctions between these configurations are outlined in the table below:
%\begin{table}[h]
%\centering
%\caption{Model Descriptions \\}
% \scalebox{0.8}{
%\resizebox{0.8\textwidth}{!}{
%\begin{tabular}{l|c|c}
%\hline
% \toprule
%& BioBERT Base & BioBERT Large \\
%\hline
% \multirow{3}{*}{cellline}
%Number of Layers & 12 & 24 \\
%Number of attention heads per layer & 12 & 16 \\
%Number of units in hidden layer & 768 & 1024 \\
%Vocabulary & Original BERT & Original BERT \\ & & + custom 30K biomedical \\
% \bottomrule
%\hline
%\end{tabular}
%}
%\label{tab:models}
%\end{table}
BioBERT-Base was used for the experiments in Table~\ref{tab:chemprot-vs-merged-results}.    
To comprehensively assess performance on our merged dataset, we also conducted experiments with a BioBERT-Large model.  Specifically, we developed three variants of the BioBERT approach:  BioBERT-Base with the top model, BioBERT-Large without the top model, and BioBERT-Large with the top model.

%Additionally, to comprehensively assess the system's performance, we evaluated it both with and without the utilization of the aforementioned top model.
%We have developed two variants of the RE model, each built upon different configurations of BioBERT, as detailed in Table~\ref{tab:models}. Additionally, to comprehensively assess the system's performance, we evaluated it both with and without the utilization of the aforementioned top model.

The $F_1$ scores obtained from the test set across five individual runs are presented in Table~\ref{tab:base-large-results}. These results consistently demonstrate a pattern: BioBERT-Large consistently outperforms BioBERT-Base in terms of $F_1$ scores. Additionally, the inclusion of the top model leads to an enhancement in the $F_1$ score. However, note that this enhancement is not statistically significant, indicating that achieving state-of-the-art results does not necessarily require the adoption of a more complex system.

\begin{table}[h]
\centering
\caption{Results for Different BioBERT-Based Models\\}
% \scalebox{0.8}{
\resizebox{0.9\textwidth}{!}{
\begin{tabular}{|l|c|c|c|c|c|c|c|c|c|}
\hline
% \top& rule
      & \multicolumn{3}{|c|}{BioBERT-Base} & \multicolumn{3}{|c|}{BioBERT-Large} & \multicolumn{3}{|c|}{BioBERT-Large} \\
      & \multicolumn{3}{|c|}{w/Top Model}  & \multicolumn{3}{|c|}{w/o Top Model} & \multicolumn{3}{|c|}{w/ Top Model} \\
\hline
      & P & R & F                          & P & R & F                         & P & R & F \\ 
\hline
CPR-1 & 0.7899	& 0.8607	& 0.8233 & 0.8773	& 0.8734	& 0.8749 & 0.8623	& 0.8871	& 0.8733\\
\hline
CPR-2 & 0.5574	& 0.4903	& 0.5202 & 0.5331	& 0.5085	& 0.5199 & 0.5380	& 0.5066	& 0.5213\\
\hline
CPR-3 & 0.8333	& 0.8677	& 0.8494 & 0.8713	& 0.8811	& 0.8760 & 0.8671	& 0.8809	& 0.8735\\
\hline
CPR-4 & 0.8697	& 0.9253	& 0.8965 & 0.9010	& 0.9310	& 0.9157 & 0.8929	& 0.9311	& 0.9116\\
\hline
CPR-5 & 0.8898	& 0.8956	& 0.8920 & 0.8802	& 0.9425	& 0.9093 & 0.8861	& 0.9319	& 0.9080\\
\hline
CPR-6 & 0.8834	& 0.9402	& 0.9105 & 0.9254	& 0.9533	& 0.9390 & 0.9276	& 0.9440	& 0.9357\\
\hline
CPR-9 & 0.7325	& 0.8279	& 0.7772 & 0.7928	& 0.8253	& 0.8067 & 0.8203	& 0.8157	& 0.8174\\
\hline
CPR-10 & 0.9292	& 0.9140	& 0.9215 & 0.9325	& 0.9256	& 0.9290 & 0.9319	& 0.9270	& 0.9294\\
\hline
\hline
Micro F1 & \multicolumn{3}{|c|}{0.8836} & \multicolumn{3}{|c|}{0.8948} & \multicolumn{3}{|c|}{0.8952} \\
Std dev  & \multicolumn{3}{|c|}{0.0021} & \multicolumn{3}{|c|}{0.0026} & \multicolumn{3}{|c|}{0.0035} \\
\hline
\end{tabular}
}
\label{tab:base-large-results}
\end{table}

\subsection{BioBERT and GCN-BERT Comparison}

In this section, we compare the performance of incorporating GCN with BioBERT. 
To ensure robustness of our results, we report them as an average of five runs. The parameters and evaluation metrics are presented in Table \ref{tab:benchmarking-results}. Table \ref{tab:comparison} presents results per class for Experiment 5 from Table \ref{tab:benchmarking-results} and compares the results with those obtained using BioBERT-Base and BioBERT-Large as described in Section~\ref{sec:biobert}. As noted in Section~\ref{sec:data}, CPR 7 and CPR 8 are not included in the results due to the low number of training data instances for those relations. The results show that incorporating the GCN does increase the overall F-1 score for some instances. 

\begin{table*}[h]
\centering
\caption{GCN-BERT Experimental Results}
\large
\resizebox{1.2\textwidth}{!}{
\renewcommand{\arraystretch}{1.5} 
\begin{tabular}{|c||l|c|c|c|c|}
\hline
Ex. \# & Parameters & Micro F1 (std dev)& Weighted avg F1 (std dev)& Macro F1
(std dev)\\
\hline
1 & Downsampling (ratio = 0.6) & 83.253 (0.007)& 83.313 (0.021)& 71.2 (0.005)\\ \hline 
2 & No downsampling & 87.348 (0.008)& 87.184 (0.003)& 69.2 (0.003)\\ \hline 
3 & Reverse entity masking without downsampling & 89.146 (0.023)& 89.057 (0.090)& 74.69 (0.004)\\ \hline 
4 & BioBERT-Large without downsampling & 88.618
(0.010)& 88.453 (0.003)& 71.78 (0.006)\\  \hline 
5 & \textbf{Reverse entity masking with Downsampling (ratio 0.6)} & \textbf{89.326 (0.012)}& \textbf{89.438 (0.017)}& \textbf{75.68 (0.002)}\\
\hline 
\end{tabular}
}
\label{tab:benchmarking-results}
\end{table*}

\begin{table*}[h]
\centering
\caption{Comparing BioBERT-Base, BioBERT-Large and GCN-BERT}
% \scalebox{0.8}{
\resizebox{0.9\textwidth}{!}{
\begin{tabular}{|l|c|c|c|c|c|c|c|c|c|}
\hline
% \top& rule
      & \multicolumn{3}{|c|}{BioBERT-Base} & \multicolumn{3}{|c|}{BioBERT-Large} & \multicolumn{3}{|c|}{GCN-BERT} \\
\hline
      & P & R & F                          & P & R & F                         & P & R & F \\ 
\hline
CPR-1 & 0.7899	& 0.8607	& 0.8233 & 0.8623	& 0.8871	& {\bf 0.8733}  & 0.6929 & 0.6730 & 0.6874 \\
\hline
CPR-2 & 0.5574	& 0.4903	& 0.5202 & 0.5380	& 0.5066	& 0.5213  & 0.7514 & 0.6298 & {\bf 0.6766}\\
\hline
CPR-3 & 0.8333	& 0.8677	& 0.8494 & 0.8671	& 0.8809	& {\bf 0.8735} & 0.7419 & 0.7657 & 0.7535\\
\hline
CPR-4 & 0.8697	& 0.9253	& 0.8965 & 0.8929	& 0.9311	& {\bf 0.9116}  & 0.8043 & 0.8322 & 0.8141\\
\hline
CPR-5 & 0.8898	& 0.8956	& 0.8920 & 0.8861	& 0.9319	& {\bf 0.9080} & 0.7238 & 0.7103 & 0.7128 \\
\hline
CPR-6 & 0.8834	& 0.9402	& 0.9105 & 0.9276	& 0.9440	& {\bf 0.9357} & 0.7378 & 0.8087 & 0.7661 \\ 
\hline
CPR-9 & 0.7325	& 0.8279	& 0.7772 & 0.8203	& 0.8157	& {\bf 0.8174} & 0.7372 & 0.6761 & 0.7053 \\ 
\hline
CPR-10 & 0.9292	& 0.9140	& 0.9215 & 0.9319	& 0.9270	& 0.9294 & 0.9326 & 0.9395 & {\bf 0.9355} \\
\hline
\hline
Micro F1 & \multicolumn{3}{|c|}{0.8836} & \multicolumn{3}{|c|}{0.8952} & \multicolumn{3}{|c|}{0.8933} \\
\hline
\end{tabular}
}
\label{tab:comparison}
\end{table*}

%\label{tab:base-large-results}
%\end{table}
%\begin{table}[ht]
%\centering
%\resizebox{0.5\textwidth}{!}{  % Increase font size
%\renewcommand{\arraystretch}{1.5}  % Increase row height
%\begin{tabular}{|l|l|l|l|}
%\hline
%Relation Class& Precision & Recall & F1-score \\
%\hline
%CPR-1 & 0.6929 (0.035)& 0.6730 (0.010)& 0.6874 (0.021)\\ \hline
%CPR-2 & 0.7514 (0.012)& 0.6298 (0.020)& 0.6766 (0.009)\\ \hline
%CPR-3 & 0.7419 (0.020)& 0.7657 (0.024)& 0.7535 (0.021)\\ \hline
%CPR-4 & 0.8043 (0.005)& 0.8322 (0.011)& 0.8141 (0.002)\\ \hline
%CPR-5 & 0.7238 (0.025)& 0.7103 (0.030)& 0.7128 (0.006)\\ \hline
%CPR-6 & 0.7378 (0.013)& 0.8087 (0.014)& 0.7661 (0.022)\\ \hline
%CPR-9 & 0.7372 (0.008)& 0.6761 (0.010)& 0.7053 (0.095)\\ \hline
%CPR-10 & 0.9326 (0.001)& 0.9395 (0.001)& 0.9355 (0.001)\\ \hline
%\end{tabular}
%}
%\caption{Results per class on Experiment 5}
%\label{tab:per-class-results}
%\end{table}

%\begin{table}[ht]
%\centering
%\resizebox{1.0\textwidth}{!}{
%%\renewcommand{\arraystretch}{1.5}  % Increase row height
%\begin{tabular}{|c|l|l|l|}
%\hline
%Results & Macro F1 & Micro F1 & Weighted F1 \\
%\hline
%GCN-BERT (Experiment 5)& 75.68 (0.002)& 89.326 (0.017)& 89.438 (0.012)\\ \hline
%BioBERT-Base & 86.06 (0.006)& 92.21 (0.004)& 92.18 (0.003)\\\hline
%BioBERT-Large & 87.14 (0.008)& 93.22 (0.004)& 93.48 (0.005)\\\hline
%\end{tabular}
%}
%\caption{Model Performance Metrics comparisons}
%\label{tab:comparison}
%\end{table}

% \subsection{Result Analysis}
Following the analysis of GCN-BERT's results, a plausible pattern can be detected in its successes and shortcomings, particularly concerning the dominant relations that entities hold within a paragraph. Consider the following text:

\textit{Some clinically used compounds, such as acetazolamide, methazolamide, ethoxzolamide, \textbf{dichlorophenamide}, \textbf{dorzolamide}, \textbf{brinzolamide}, \textbf{topiramate}, sulpiride, and indisulam, or the orphan drug benzolamide, showed effective \textbf{hCA VI} inhibitory activity, with inhibition constants of 0.8-79 nM. The best inhibitors were brinzolamide and sulpiride (KI values of 0.8-0.9 nM), the latter compound being also a CA VI-selective inhibitor}.

% Correctly predicted relations:
% \begin{table}
%     \centering
%     \begin{tabular}{|c|c|c|} \hline 
%          Entity 1&  Entity 2& Relation\\ \hline 
%          dichlorophenamide&  hCA VI& CPR-4\\ \hline 
%          brinzolamide&  hCA VI& CPR-4\\ \hline
%  topiramate& hCA VI&CPR-4\\\hline
%     \end{tabular}
%     \caption{Caption}
%     \label{tab:my_label}
% \end{table}

% Incorrect predictions:

Here, the entity "hCA VI" is related to most of the other entities in the paragraph, namely, dichlorophenamide, dorzolamide, brinzolamide, and topiramate, under the "CPR-4" relation. The model is able to correctly classify these relations since "CPR-4" is well-represented in the training data. However, the model encounters difficulty in cases where "hCA VI" has a relation other than "CPR-4". Additionally, since the dataset is imbalanced, with many samples having a "NOT" relation, there are many false negatives in the model's predictions.

In summary, the performance of GCN-BERT appears to hinge on the frequency and dominance of specific relations associated with entities in a given paragraph. While the model excels in cases where an entity has the same type of relation with other entities, it faces challenges when confronted with less frequent or alternative relations, leading to misclassifications in certain instances.

\section{Discussion}
\label{sec:discussion}

In this section, we discuss the our analysis of the results reported in Section~\ref{sec:results}.

\subsection{Effect of Class Sample Size} 
In our analysis of the experiment results, we observed an expected relationship between the number of training samples in a class and the model's performance. Specifically, the model consistently performed better on classes that had a higher number of samples in the training data.

For instance, the 'NO RELATIONS' class (CPR-10) consistently had the highest number of samples, and as a result, the model consistently delivered the best performance on this class. This trend was also observed in the case of CPR-4, which had the second-highest number of samples and consistently yielded the second-best performance across different tuning techniques.

In Experiments 1 and 2 in Table~\ref{tab:benchmarking-results}, we compared the results of our model with and without downsampling of the negative samples. We found that, without downsampling in Experiment 2, the model achieved a higher micro F1 score compared to Experiment 1, which involved downsampling. The absence of downsampling in Experiment 2 led to more samples in the CPR-10 class, resulting in better performance in this specific class. However, due to the imbalance in the dataset, Experiment 2 performed worse than Experiment 1 in other classes. This discrepancy is evident when looking at the macro F1 score, which reflects the overall performance across all classes.

\subsection{Handling class imbalance} Given the presence of an imbalanced dataset with numerous negative samples, we employed specific techniques to address this challenge, including downsampling and weighted loss. Notably, our findings indicate that a downsampling ratio of 0.6 offers an optimal balance for handling the data imbalance. This strategy not only aids in mitigating class imbalance issues but also enables the model to allocate sufficient attention to various relation classes beyond the negative samples.

\subsection{Entity Masking} 
Our experiments confirm the effectiveness of masking target entities with their corresponding semantic types, leading to a significant 6\% increase in accuracy compared to the alternative approach. This result implies that the presence of other entities within the sentence plays a substantial role in enhancing the model's contextual understanding, thereby improving its ability to predict relations among entities. 

\subsection{Graph Embeddings} 
While the vocabulary graph embeddings contributes to the GCN-BERT's ability to capture the global context, it is worth noting that the vocabulary graph is constructed using all words within the corpus. This approach, though informative, may introduce a potential challenge for the model. The graph's inclusion of nodes for all words in the corpus provides an abundance of information, which might inadvertently introduce noise, diverting the model's focus from its primary task of capturing relationships between chemical entities. One possible improvement could involve constructing the vocabulary graph using only entity words.
%, as discussed in Section \ref{sec:filtered_vocab}. This noise may contribute to its lower performance compared to BioBERT on individual classes, as shown in Table \ref{tab:comparison}.

\subsection{Difficulty handling unseen words during inference}
One potential drawback of the GCN-BERT approach is its reliance on a vocabulary graph constructed using words from both the training and test corpora. This design choice might hinder the model's performance on unseen data containing words that were not part of the original corpus. Specifically, generating graph embeddings for such words during inference can be challenging because nodes corresponding to those words are absent from the vocabulary graph.

\section{Future Work}
\label{sec:future-work}

First, given the performance gains observed from merging the ChemProt and DrugProt datasets, we plan to explore additional data sources and methods for combining them to create even larger, more extensive datasets across a more diverse set of relationship types.

Second, based on our observations, we propose that the below avenues will help further enhance the performance of both the BioBERT and GCN-BERT models. Firstly, better handling of OOV words. The vocabulary graph contains nodes, which are word tokens tokenized by the BioBERT tokenizer. It splits OOV words into subwords and uses '\#\#' to indicate that they are part of a larger word. For example, the word 'xyz' may be tokenized as 'x' and '\#\#yz.' This issue is extremely apparent with chemical tokens. We believe that incorporating even more domain-specific knowledge into the text corpus to generate these embeddings and tokens could enhance the representation of chemical and gene entities within the vocabulary graph. An example of such an approach is \cite{zhang2019biowordvec}, which uses a subword embedding model and incorporates domain knowledge from MeSH. Specifically for GCN-BERT, Since the vocabulary graph is constructed using word tokens as nodes, the adoption of a domain-specific tokenization scheme or the inclusion of vocabulary words in the text corpus could lead to more coherent graph embeddings.

Third, for GCN-BERT utilizing a filtered vocabulary graph. The vocabulary graph is typically constructed using all words in the vocabulary, which can introduce noise into the model. To address this, we plan to implement a "Filtered Vocabulary Graph" that contains only entities as nodes. This approach is inspired by the work of \cite{zeng2020double}, who constructed a mention graph that included both entity mention and document nodes and applied Graph Convolutional Networks to transform it into an entity graph. Focusing more on entities in the vocabulary graph could yield more accurate and informative results for our model.

Lastly, separate the training of the GCN model and BERT. As suggested by previous authors \cite{mahendran2022graph}, we plan to explore the possibility of training the Graph Convolutional Network and BERT separately. Subsequently, we will combine the embeddings before passing them through the final classification layer. Training them separately might enhance the learning capabilities of both models, leading to improved representations and, consequently, enhanced model performance.

\section{Conclusion}

This paper presents a new, larger, more comprehensive data set created by merging the ChemProt and DrugProt datasets to augment sample counts and improve model accuracy. We evaluated the merged data set using two relationship extraction models: BioBERT and GCN-BERT. The BioBERT architecture aims to capture contextualized information contained within the sentence, while the GCN-BERT architecture aims to classify relations by leveraging both global information through graph embeddings and local contextual information through BERT. We demonstrated significant improvements in model performance, particularly in CPR groups shared between the datasets.

 %After rigorous hyperparameter tuning, we achieved an accuracy of 89.33\% with the use of reverse entity masking and downsampling. While GCN-BERT demonstrates good overall accuracy compared to state-of-the-art models, it falls short in individual class performance. We identify potential reasons for this underperformance and propose techniques such as a filtered vocabulary graph and improved handling of out-of-vocabulary (OOV) words, which could enhance the model's performance for individual classes. We also highlight the factors that contributed to the model's improved performance, including reverse entity masking, weighted loss, and downsampling. 

\section{Data and Software Availability}
The data and software are publicly available at: https://github.com/bmcinnes/UCSD-VCU-IE-Collaboration

%\section*{References}
\section*{Acknowledgement}\label{sec:acknowledgement}
    This work was funded in part by NSF award numbers 1939885 and 1939951, and utilized computer resources provided in part by NSF award numbers 1730158, 2100237, 2120019 and 2316003.

\bibliography{citations}
\end{document}